\documentclass[letterpaper, 10 pt, conference]{ieeeconf}  

\usepackage{setspace}
\usepackage{multirow}
\usepackage{booktabs}
\usepackage{verbatim} 
\usepackage{empheq}
\usepackage[normalem]{ulem}
\usepackage{soul}
\usepackage{xcolor}
\usepackage{algorithm,algorithmic}
\usepackage{bbm}
\usepackage{amsmath,amssymb}
\usepackage{amsfonts}
\usepackage{graphicx}
\graphicspath{{./figures/}}
\usepackage{subcaption}
\usepackage{pict2e,color}
\usepackage{cite}

\newtheorem{definition}{Definition}
\newtheorem{theorem}{Theorem}

\IEEEoverridecommandlockouts                              

\usepackage{tikz}
\usepackage{textcomp}
\usepackage{hyperref}
\usepackage{lipsum}

\newcommand\copyrighttext{%
  \footnotesize \textcopyright 2012 IEEE. Personal use of this material is permitted.
  Permission from IEEE must be obtained for all other uses, in any current or future
  media, including reprinting/republishing this material for advertising or promotional
  purposes, creating new collective works, for resale or redistribution to servers or
  lists, or reuse of any copyrighted component of this work in other works.
  \href{https://ieeexplore.ieee.org/document/9992450}{DOI 10.1109/CDC51059.2022.9992450}}
\newcommand\copyrightnotice{%
\begin{tikzpicture}[remember picture,overlay]
\node[anchor=south,yshift=10pt] at (current page.south) {\fbox{\parbox{\dimexpr\textwidth-\fboxsep-\fboxrule\relax}{\copyrighttext}}};
\end{tikzpicture}%
}

\title{\LARGE \bf
Risk-Averse Reinforcement Learning\\
via Dynamic Time-Consistent Risk Measures
}

\author{Xian Yu$^{1}$ and Siqian Shen$^{2}$
\thanks{$^{1}$Xian Yu is with the Department of Integrated Systems Engineering,
        The Ohio State University, Columbus, Ohio, USA.
        {\tt\small yu.3610@osu.edu}.}%
\thanks{$^{2}$Siqian Shen is with the Department of Industrial and Operations Engineering,
        University of Michigan, Ann Arbor, Michigan, USA. {\tt\small  siqian@umich.edu}.
}
}

\begin{document}

\maketitle
\copyrightnotice
\thispagestyle{empty}
\pagestyle{empty}

\begin{abstract}
Traditional reinforcement learning (RL) aims to maximize the expected total reward, while the risk of uncertain outcomes needs to be controlled to ensure reliable performance in a \textit{risk-averse} setting. In this paper, we consider the problem of maximizing dynamic risk of a sequence of rewards in infinite-horizon Markov Decision Processes (MDPs). We adapt the Expected Conditional Risk Measures (ECRMs) to the infinite-horizon risk-averse MDP and prove its time consistency. Using a convex combination of expectation and conditional value-at-risk (CVaR) as a special one-step conditional risk measure, we reformulate the risk-averse MDP as a risk-neutral counterpart with augmented action space and manipulation on the immediate rewards. We further prove that the related Bellman operator is a contraction mapping, which guarantees the convergence of any value-based RL algorithms. Accordingly, we develop a risk-averse deep Q-learning framework, and our numerical studies based on two simple MDPs show that the risk-averse setting can reduce the variance and enhance robustness of the results.
\end{abstract}

\section{INTRODUCTION}
Sequential decision making is of vital importance to the design and operations of real-world complex systems in transportation, energy and healthcare applications, where agents interact with the uncertain environment and make decisions over time to maximize certain forms of cumulative reward. The related dynamic information gathering and decision-making problems can be modeled using Markov Decision Processes (MDPs) \cite{puterman2014markov}. Recently, reinforcement learning (RL) becomes popular for solving MDPs. In standard RL, one seeks a policy that maximizes the expected total discounted rewards, which we refer to as \textit{risk-neutral} RL. However, maximizing the expected reward does not necessarily avoid the maybe rare occurrences of undesirable outcomes, and in a situation where it is important to maintain reliable performance, we aim to evaluate and control the \textit{risk}. 

There mainly exist two types of uncertainties involved in a system: (i) parameter uncertainty due to errors when estimating the transition probabilities, rewards, etc., and (ii) inherent uncertainty due to stochastic transitions of the environment. The literature in robust MDP mainly focus on tackling parameter uncertainties \cite{nilim2005robust, iyengar2005robust, wiesemann2013robust, lim2013reinforcement, xu2010distributionally, yu2015distributionally, nakao2021distributionally}. In this paper, we consider the risk caused by the inherent uncertainty of the stochastic environment, and we aim to maximize certain risk measures of the stochastic reward.  

Researchers have developed different ways to incorporate risk components into MDPs, where the objectives are to maximize the worst-case outcome \cite{heger1994consideration, coraluppi2000mixed} or to control/minimize some notion of risk, e.g., reward variance \cite{howard1972risk, markowitz2000mean, borkar2002q, tamar2012policy, la2013actor}. 
In this paper, we employ coherent risk measures \cite{artzner1999coherent}, including the conditional value-at-risk (CVaR) \cite{rockafellar2000optimization}, also known as the expected shortfall, to quantify the amount of tail risk. We refer the interested readers to \cite{rockafellar2002conditional, ruszczynski2006optimization, shapiro2009lectures} for in-depth discussions about CVaR modeling and solution approaches. 

In a sequential decision making framework, the risk can be measured on the total discounted reward or in a nested way, leading to static and dynamic risk measures. In the former, the total discounted reward is modeled by a single random variable, and measured by a one-shot risk metric, without considering the temporal relationship between the risk in different stages. Ref.\ \cite{yu2017dynamic} applied static risk measures on the total discounted cost and proposed grid search and convex approximation schemes to solve the resultant risk-aware MDPs. Ref.\ \cite{chow2014algorithms,tamar2015optimizing} considered static CVaR and developed methods for computing the gradient of CVaR. 
However, a drawback of static risk measure is that it does not satisfy time consistency, an important property that ensures consistent risk preferences over time. Ref.\ \cite{rudloff2014time} provided an example of a multi-period portfolio selection problem where a static risk measure leads to inconsistent risk preferences in an irrational fashion and specifically drives a risk-averse decision maker to show risk-neutral preferences at intermediate stages. 

A commonly-used and well-established definition of time consistency in dynamic risk measures is as follows \cite{ruszczynski2010risk}: if a certain reward is considered less risky at stage $k$, then it should also be considered less risky at an earlier stage $l<k$. In \cite{ruszczynski2010risk}, the authors proved that a risk measure is time consistent \emph{iff} it can be represented as a composition of one-step conditional risk mappings. They also devised risk-averse value iteration and policy iteration methods. Ref.\ \cite{chow2013stochastic,chow2014framework} studied dynamic, time-consistent risk measures either in the objective function or in the constraints. However, all of these methods employed model-based algorithms, which rely on the model of transition probabilities and rewards. Our approach differs from the existing methods by recasting the risk-averse MDPs as the traditional risk-neutral MDPs with augmented action space and revised immediate rewards, which can be solved by any standard RL approaches. 

The main contributions of this paper are as follows. We adopt the definition in \cite{ruszczynski2010risk} and apply the expected conditional risk measures (ECRMs) \cite{homem2016risk} on infinite-horizon MDPs. We show that ECRMs are time-consistent, and they can be recast in a nested way. Another appealing property is that MDPs with ECRMs can be recast as risk-neutral MDPs with manipulation on the immediate reward and augmented action space. We show that the risk-averse Bellman operator is a contraction mapping, which ensures the convergence of any value-based RL algorithms. To the best of our knowledge, this paper also proposes the first model-free risk-averse RL approach based on deep Q-learning framework using dynamic time-consistent risk measures. We finally test and illustrate our approaches using a 10-state MDP and a 7-state random walk environment \cite{sutton1988learning,sutton2018reinforcement}.

\section{Preliminaries}\label{sec:background}
\subsection{Classical Markov Decision Processes}
We consider a discrete-time MDP that can be represented by the tuple: $(\mathcal{S},\mathcal{A},P, P_r,r, \gamma, s_1)$, where $\mathcal{S}$ is a finite set of states, $\mathcal{A}$ is a finite set of actions,
$P:\mathcal{S}\times\mathcal{A}\times\mathcal{S}\to [0,1]$ is the state transition probability, respectively. An immediate reward $r\in \mathbb{R}$ is distributed according to the reward probability $P_r: \mathbb{R}\times \mathcal{S}\times\mathcal{A}\times\mathcal{S}\to [0,1]$. Without loss of generality, we assume a fixed initial state $s_1$ and consider an infinite-horizon MDP with $\gamma$ being the discount factor. The goal of traditional RL is to find a deterministic stationary policy $\pi: \mathcal{S}\to\mathcal{A}$ that maximizes the Q-function which is the expected accumulated reward:
\begin{align*}
    Q^{\pi}(s, a) &= \mathbb{E}\left[\sum_{t=1}^{\infty}\gamma^{t-1} r(s_t,a_t)\right], ~ \nonumber\\
    &s_t\sim P(\cdot|s_{t-1},a_{t-1}),\  a_t=\pi(s_t),\ s_1 = s,\ a_1 = a.
\end{align*}

The action value function $Q^{\pi}(s,a)$ satisfies the well-known Bellman's equation \cite{bellman1966dynamic}:
\begin{align*}
    Q^{\pi}(s_t,a_t)& = \mathbb{E}[r(s_t,a_t)] + \gamma \mathbb{E}[Q^{\pi}(s_{t+1}, a_{t+1})],\nonumber\\
    &s_{t+1}\sim P(\cdot|s_t, a_t),\ a_{t+1}=\pi(s_{t+1}).
\end{align*}
For the ease of presentation, we rewrite $r(s_t,a_t)$ as $r_t$ for all $t\ge 1$ and denote vector $(a_1,\ldots,a_t)$ as $a_{[1,t]}$ in the rest of this paper.
\subsection{Coherent One-Step Conditional Risk Measures}
\label{sec:def-coherent}

Before formally introducing dynamic risk measures, we first define coherent one-step conditional risk measures. Consider a probability space $(\Xi,\mathcal{F},P)$, and let $\mathcal{F}_1\subset\mathcal{F}_2\subset\ldots$ be sub-sigma-algebras of $\mathcal{F}$ such that each $\mathcal{F}_t$ corresponds to the information available up to (and including) stage $t$, with $Z_t,\ t=1,2\ldots$ being an adapted sequence of random variables. In this paper, we interpret the variables $Z_t$ as immediate rewards. We assume that $\mathcal{F}_1=\{\emptyset,\Xi\}$, and thus $Z_1$ is in fact deterministic. Let $\mathcal{Z}_t$ denote a space of $\mathcal{F}_t$-measurable functions from $\Xi$ to $\mathbb{R}$. According to \cite{artzner1999coherent}, we have the definition for a coherent one-step conditional risk measure stated below.

\begin{definition}
A coherent one-step conditional risk measure is a mapping $\rho: \mathcal{Z}_{k+1}\to\mathcal{Z}_k$ if it satisfies the following four properties:
\begin{enumerate}
\item Monotonicity: If $Z_1,Z_2\in\mathcal{Z}_{k+1}$ and $Z_1\ge Z_2$, then $\rho(Z_1)\ge\rho(Z_2)$.
\item Convexity: $\rho(\gamma Z_1+(1-\gamma)Z_2)\le \gamma\rho(Z_1)+(1-\gamma)\rho(Z_2)$ for all $Z_1,\ Z_2\in\mathcal{Z}_{k+1}$ and all $\gamma\in[0, 1]$.
\item Translation invariance: If $W\in\mathcal{Z}_k$ and $Z\in\mathcal{Z}_{k+1}$, then $\rho(Z+W)=\rho(Z)+W$.
\item Positive Homogeneity: If $\gamma\ge 0$ and $Z\in\mathcal{Z}_{k+1}$, then $\rho(\gamma Z)=\gamma\rho(Z)$.
\end{enumerate}
In this paper, the inequalities between random variables are understood in the almost sure sense (i.e., $Z_1\ge Z_2$ if and only if $Z_1(\xi)\ge Z_2(\xi)$ for a.e. $\xi\in\Xi$.)
\end{definition}

For our problem, we consider a special class of coherent one-step conditional risk measures $\rho_t^{s_{[1,t-1]}}$ mapping from $\mathcal{Z}_t$ to $\mathcal{Z}_{t-1}$, which is a convex combination of conditional expectation and Conditional Value-at-Risk (CVaR), i.e.,
\begin{equation}
\rho_t^{s_{[1,t-1]}}(r_t)=(1-\lambda_t)\mathbb{E}[r_t|s_{[1,t-1]}]+\lambda_t \text{CVaR}_{\alpha_t}[r_t|s_{[1,t-1]}],
\label{eq:rho}
\end{equation}
where $\lambda_t\in[0,1]$ is a weight parameter to balance the expected reward and tail risk, and $\alpha_t\in(0,1)$ represents the confidence level. The expectation in \eqref{eq:rho} involves the transition from $s_{t-1}$ to $s_t$ as well as the reward distribution of $r_t$ itself. Notice that this risk measure is more general than CVaR and it has CVaR as a special case when $\lambda_t=1$. 

Following the results by \cite{rockafellar2002conditional}, the lower-tail CVaR can be expressed as the optimization problem below:
\begin{equation}
\text{CVaR}_{\alpha_t}[r_t|s_{[1,t-1]}]:=\max_{\eta_t\in\mathbb{R}}\left\lbrace \eta_t-\frac{1}{\alpha_t}\mathbb{E}[[\eta_t-r_t]_{+}|s_{[1,t-1]}]\right\rbrace,\label{eq:cvar}
\end{equation}
where $[a]_{+}:=\max\{a,0\}$, and $\eta_t$ is an auxiliary variable.  Note that the goal is to maximize the total reward (i.e., the higher the better) and this particular version of CVaR focuses on the lower tail of the distribution. As a result, maximizing the lower-tail CVaR will be risk-averse and will lead to robust solutions. 


\subsection{Dynamic Time-Consistent Risk Measures}
\label{sec:def-time}
As quoted from \cite{ruszczynski2010risk},
``\textit{The fundamental question in the field of dynamic risk measures is the following: how do we evaluate the risk of the subsequence $Z_t,\ldots, Z_T$ from the perspective of stage $t$?}''
This motivates the definitions of dynamic time-consistent risk measures.

We define $\mathcal{Z}_{t,\infty}:=\mathcal{Z}_t\times\mathcal{Z}_{t+1}\times\cdots$ for all $t\ge 1$. A sequence $Z_{[1,\infty]}\in\mathcal{Z}_{1,\infty}$ is almost surely bounded, if $\max_t \text{essup} |Z_t(\xi)|<\infty$. We adapt the definitions in Section 3 of \cite{ruszczynski2010risk} to infinite horizons as follows.
\begin{definition}
A dynamic risk measure is a sequence of monotone one-step conditional risk measures $\mathbb{F}_{k,\infty}: \mathcal{Z}_{k,\infty}\to\mathcal{Z}_k,\ k\ge 1$.
\end{definition}

\begin{definition}\label{def:time-consistent}
A dynamic risk measure $\{\mathbb{F}_{k,\infty}\}_{k=1}^{\infty}$ is called time consistent if, for all $1\le l<k$ and all sequences $Z_{[l,\infty]}, \ W_{[l,\infty]}\in \mathcal{Z}_{l, \infty}$, the conditions
\begin{align*}
    &Z_i = W_i,\ \forall i=l,\ldots, k-1, \text{and}\\
    &\mathbb{F}_{k,\infty}(Z_{[k,\infty]})\ge \mathbb{F}_{k,\infty}(W_{[k,\infty]})
\end{align*}
imply that $
    \mathbb{F}_{l,\infty}(Z_{[l,\infty]})\ge 
     \mathbb{F}_{l,\infty}(W_{[l,\infty]}).$
\end{definition}
In other words, if $Z$ will be at least as good as $W$ from the perspective of some future time $k$, then $Z$ should not be worse than $W$ from current period $l$'s perspective.  

\section{Risk-Averse RL via Dynamic Time-Consistent Risk Measures}
Our goal is to find a dynamic risk measure that is both time consistent and easy to compute. 
Given the initial state $s_1$, for any almost surely bounded sequence of immediate rewards $r_{[1,\infty]}\in\mathcal{Z}_{1,\infty}$, we consider a class of multi-period risk function $\mathbb{F}$ mapping from $\mathcal{Z}_{1,\infty}$ to $\mathbb{R}$ defined as:
\small
\begin{align}
\mathbb{F}(r_{[1,\infty]}|s_1) = &r_1+\gamma\rho^{s_1}_2(r_2)\nonumber\\
&+\lim_{T\to\infty}\sum_{t=3}^T\gamma^{t-1}\mathbb{E}_{s_{[1,t-1]}}\left[{\rho_t^{s_{[1,t-1]}}}(r_t)\right]\label{ECRMs}, 
\end{align}\normalsize
where ${\rho_t^{s_{[1,t-1]}}}$ is the coherent one-step {conditional} risk measure mapping from $\mathcal{Z}_t$ to $\mathcal{Z}_{t-1}$ defined in Eq.\ \eqref{eq:rho} to represent the risk given the information available up to (including) stage $t-1$, and the expectation is taken with respect to the random history $s_{[1,t-1]}$. We assume that the immediate reward $r_t=r(s_t,a_t)$ is deterministic given $s_t, a_t$, in order to satisfy the assumption in Sections \ref{sec:def-coherent} (i.e., $Z_1$ is deterministic). 

This class of multi-period risk measures is called expected conditional risk measures (ECRMs), first introduced in \cite{homem2016risk}. 
The authors proved several appealing properties of ECRMs from the algorithmic viewpoint. First, the risk function can be written in a nested form. Second, any risk-averse problems defined with expected CVaR is equivalent to a risk-neutral model with an augmented action space. The authors also proved that the ECRMs with finite time horizons and $\gamma = 1$ are time consistent under inherited optimality property.
We adapt the risk measures to infinite-horizon MDPs with discount factor $0<\gamma<1$ and prove the time consistency of ECRMs according to Definition \ref{def:time-consistent} in Theorem \ref{thm:ECRM-discount}.

First, we show that the risk function \eqref{ECRMs} is well defined for all almost surely bounded sequence $r_{[1,\infty]}$.
Using the tower property of expectations and the translation-invariant and positive homogeneity properties of each $\rho_t^{s_{[1,t-1]}}$, we rewrite $\mathbb{F}$ in Eq. \eqref{ECRMs} as
\small
\begin{align}
&r_1+\rho^{s_1}_2\Big(\gamma r_2+\mathbb{E}_{s_{2}}\circ{\rho_3^{s_{[1,2]}}}\Big(\gamma^2 r_3+\mathbb{E}_{s_3}^{s_{[1,2]}}\circ{\rho_4^{s_{[1,3]}}}\Big(\gamma^3 r_4\nonumber\\
&+\cdots+\mathbb{E}_{s_{t-1}}^{s_{[1,t-2]}}\circ{\rho_{t}^{s_{[1,t-1]}}}\Big(\gamma^{t-1}r_{t}+\cdots\Big)\Big)\Big)\Big),\label{eq:ECRM-nested}
\end{align}
\normalsize
where $\mathbb{E}_{s_{t-1}}^{s_{[1,t-2]}}=\mathbb{E}_{s_{t-1}}[\cdot|{s_{[1,t-2]}}]$ is the conditional expectation.
To simplify the notation, we define $\tilde{\rho}_t^{s_{[1,t-2]}}: = \mathbb{E}_{s_{t-1}}^{s_{[1,t-2]}}\circ \rho_t^{s_{[1,t-1]}}$, which maps from $\mathcal{Z}_t$ to $\mathcal{Z}_{t-2}$. Then, the multi-period risk function $\mathbb{F}$ can be recast as
\small
\begin{align}
&r_1+\rho^{s_1}_2\Big(\gamma r_2+\tilde{\rho}_3^{s_{[1]}}\Big(\gamma^2r_3+\tilde{\rho}_4^{s_{[1,2]}}\Big(\gamma^3r_4\nonumber\\
&+\cdots+\tilde{\rho}_t^{s_{[1,t-2]}}\Big(\gamma^{t-1}r_{t}+\cdots\Big)\Big)\Big)\Big).\label{eq:ECRM}
\end{align}\normalsize
This gives us a formulation similar to Eq.\  (26) in Section 6 of \cite{ruszczynski2010risk}. The only difference is that their one-step conditional risk measures are mappings from $\mathcal{Z}_t$ to $\mathcal{Z}_{t-1}$, but our $\tilde{\rho}_t^{s_{[1,t-2]}}$ is mapping from $\mathcal{Z}_t$ to $\mathcal{Z}_{t-2}$. Because $\mathcal{F}_{t-2}\subset \mathcal{F}_{t-1}$, any $\mathcal{F}_{t-2}$-measurable function from $\Xi$ to $\mathbb{R}$ is also $\mathcal{F}_{t-1}$-measurable (i.e., $\mathcal{Z}_{t-2}\subset\mathcal{Z}_{t-1}$). In other words, $\tilde{\rho}_t^{s_{[1,t-2]}}$ can be also regarded as a mapping from $\mathcal{Z}_t$ to $\mathcal{Z}_{t-1}$. According to Theorem 3 in \cite{ruszczynski2010risk}, the multi-period risk measure $\mathbb{F}(r_{[1,\infty]}|s_1)$ is well defined on the set of almost surely bounded sequences $r_{[1,\infty]}\in\mathcal{Z}_{1,\infty}$, i.e., the limit in \eqref{ECRMs} exists.

Next, we define a dynamic risk measure $\{\mathbb{F}_{k,\infty}\}_{k=1}^{\infty}$ as $\mathbb{F}_{1,\infty}(r_{[1,\infty]})= \mathbb{F}(r_{[1,\infty]}|s_1)$,
and for $k\ge 2$,
    \begin{align}
    \mathbb{F}_{k,\infty}(r_{[k,\infty]})=&\gamma^{k-1}r_k+\tilde{\rho}_{k+1}^{s_{[1,k-1]}}\Big(\gamma^kr_{k+1}\nonumber\\
    &+\tilde{\rho}_{k+2}^{s_{[1,k]}}\Big(\gamma^{k+1}r_{k+2}+\cdots\Big)\Big).\label{eq:dynamic_k}
\end{align}\normalsize

\begin{theorem}
\label{thm:ECRM-discount}
Consider the dynamic risk measures in \eqref{eq:dynamic_k} with an infinite time horizon, where each $\rho_t^{s_{[1,t-1]}}$ is a coherent one-step conditional risk measure and $0<\gamma<1$. Then, $\{\mathbb{F}_{k,\infty}\}_{k=1}^{\infty}$ is time consistent.
\end{theorem}

\proof
We first check the conditions in Definition \ref{def:time-consistent} when $l=1$.
Given \eqref{eq:ECRM} and \eqref{eq:dynamic_k}, we have for all $k\ge 3$,
\small
\begin{align*}
    & \mathbb{F}_{1,\infty}({r}_{[1,\infty]}) = r_1 + \rho_2^{s_1}\Big(\mathbb{F}_{2,\infty}({r}_{[2,\infty]})\Big)\\
    &= r_1 + \rho_2^{s_1}\Big(\gamma r_2 + \tilde{\rho}_3^{s_{1}}\Big(\gamma^2r_3+\ldots+ \tilde{\rho}_k^{s_{[1,k-2]}}\Big(\mathbb{F}_{k,\infty}({r}_{[k,\infty]})\Big)\Big)\Big).
    \end{align*}
    \normalsize
If $\mathbb{F}_{k,\infty}({r}_{[k,\infty]})\ge \mathbb{F}_{k,\infty}({u}_{[k,\infty]})$ and $r_i=u_i,\ \forall i=1,\ldots, k-1$, we have
$\mathbb{F}_{1,\infty}({r}_{[1,\infty]})\ge \mathbb{F}_{1,\infty}({u}_{[1,\infty]})$ because of the monotonicity of $\rho_2^{s_1}$ and $\tilde{\rho}_t^{s_{[1,t-2]}},\ \forall t\ge 3$.

Now for $l\ge 2$, we have
\small
    \begin{align*}
    \mathbb{F}_{l,\infty}({r}_{[l,\infty]}) =&\gamma^{l-1}r_l + \tilde{\rho}_{l+1}^{s_{[1,l-1]}}\Big(\gamma^lr_{l+1} +\ldots\\ &+\tilde{\rho}_k^{s_{[1,k-2]}}\Big(\mathbb{F}_{[k,\infty]}({r}_{[k,\infty]})\Big)\Big),\ \forall k\ge l+1.
\end{align*}\normalsize
If $\mathbb{F}_{k,\infty}({r}_{[k,\infty]})\ge \mathbb{F}_{k,\infty}({u}_{[k,\infty]})$ and $r_i=u_i,\ \forall i=l,\ldots, k-1$, we have
$\mathbb{F}_{l,\infty}({r}_{[l,\infty]})\ge \mathbb{F}_{l,\infty}({u}_{[l,\infty]})$ because of the monotonicity of $\tilde{\rho}_t^{s_{[1,t-2]}},\ \forall t\ge {l+1}$.
This completes the proof.
\endproof

Using the specific risk measure defined in \eqref{eq:rho} and combining with equations \eqref{eq:cvar} and \eqref{eq:ECRM-nested},  we have
\footnotesize
\begin{align}
&\max_{a_{[1,\infty]}}\mathbb{F}(r_{[1,\infty]}|s_1)\nonumber\\
    =&\max_{a_1,\eta_2}\Big\{r(s_1,a_1)+\gamma\lambda_2\eta_2+\gamma\mathbb{E}^{s_1}_{s_2}\Big[\max_{a_2,\eta_3}\Big\{-\frac{\lambda_2}{\alpha_2}[\eta_2-r(s_2,a_2)]_+\nonumber\\
    &+(1-\lambda_2)r(s_2,a_2)+\gamma\lambda_3\eta_3+\gamma\mathbb{E}^{s_2}_{s_3}\Big[\max_{a_3,\eta_4}\Big\{-\frac{\lambda_3}{\alpha_3}[\eta_3-r(s_3,a_3)]_+\nonumber\\
    &+(1-\lambda_3)r(s_3,a_3)+\gamma\lambda_4\eta_4+\cdots\Big\}\Big]\Big\}\Big]\Big\},\label{eq:nested2}
\end{align}\normalsize
where we apply the Markov property to recast $\mathbb{E}_{s_{t-1}}^{s_{[1,t-2]}}$ as $\mathbb{E}_{s_{t-1}}^{s_{t-2}}$. The auxiliary variable $\eta_t$ from Eq.\ \eqref{eq:cvar} is decided before taking conditional expectation $\mathbb{E}^{s_{t-1}}_{s_t}$ and thus it can be regarded as a $(t-1)$-stage action, similar to $a_{t-1}$. From here, we observe that the only differences between \eqref{eq:nested2} and traditional RL are the augmentation of the action space to include auxiliary action $\eta_t$ to help learn the distribution of returns, and the manipulation on immediate rewards (i.e., replacing $r(s_t,a_t)$ with $-\frac{\lambda_t}{\alpha_t}[\eta_t-r(s_t,a_t)]_++(1-\lambda_t)r(s_t,a_t)+\gamma\lambda_{t+1}\eta_{t+1}$). Moreover, if we set $\lambda_t=0,\ \forall t\ge 2$, the formulation \eqref{eq:nested2} will reduce to the risk-neutral RL problem.

This leads to the following definition of action value functions and risk-averse Bellman operator. We prove that the risk-averse Bellman operator is a contraction mapping in Theorem \ref{prop:contraction}.
\begin{definition} 
A risk-averse Bellman operator is given by
\small
\begin{align}
&\mathcal{T}Q(s_t,{\color{black}\eta_t},a_t,\eta_{t+1}) = -\frac{\lambda_t}{\alpha_t}[\eta_t-r(s_t,a_t)]_+ + (1-\lambda_t)r(s_t,a_t)\nonumber\\
&+\gamma\lambda_{t+1}\eta_{t+1}+ \gamma \mathbb{E}^{s_t}_{s_{t+1}}[\max_{a_{t+1},\eta_{t+2}} Q(s_{t+1},{\color{black}\eta_{t+1}},a_{t+1},\eta_{t+2})]. \label{eq:risk-averse-bellman}
\end{align}\normalsize
\end{definition}

\begin{theorem}
\label{prop:contraction}
The risk-averse Bellman operator \eqref{eq:risk-averse-bellman} is a contraction mapping.
\end{theorem}
\proof
We have
\small
\begin{align*}
    &||\mathcal{T}Q_1(s_t, {\color{black}\eta_t},a_t,\eta_{t+1}) - \mathcal{T}Q_2(s_t, {\color{black}\eta_t},a_t,\eta_{t+1})||_{\infty}\\
    = &\max_{s_t, {\color{black}\eta_{t}},a_t,\eta_{t+1}} |\mathcal{T}Q_1(s_t, {\color{black}\eta_t},a_t,\eta_{t+1}) - \mathcal{T}Q_2(s_t, {\color{black}\eta_t},a_t,\eta_{t+1})|\\
    \le & \max_{s_t,a_t,\eta_{t+1}}\gamma \sum_{s_{t+1}}P(s_{t+1}|s_t,a_t)|\underset{a_{t+1},\eta_{t+2}}{\max}Q_1(s_{t+1}, {\color{black}\eta_{t+1}},a_{t+1},\eta_{t+2})\\
    &-\underset{a_{t+1},\eta_{t+2}}{\max}Q_2(s_{t+1}, {\color{black}\eta_{t+1}},a_{t+1},\eta_{t+2})|\\
    {\le}  &\max_{s_t,a_t,\eta_{t+1}}\gamma \sum_{s_{t+1}}P(s_{t+1}|s_t,a_t)\max_{a_{t+1},\eta_{t+2}}|Q_1(s_{t+1}, {\color{black}\eta_{t+1}},a_{t+1},\eta_{t+2})\\
    &-Q_2(s_{t+1}, {\color{black}\eta_{t+1}},a_{t+1},\eta_{t+2})|\\
    \le &\max_{s_t,a_t,\eta_{t+1}}\gamma \sum_{s_{t+1}}P(s_{t+1}|s_t,a_t)||Q_1(s_{t+1}, {\color{black}\eta_{t+1}},a_{t+1},\eta_{t+2})\\
    &-Q_2(s_{t+1}, {\color{black}\eta_{t+1}},a_{t+1},\eta_{t+2})||_{\infty}\\
    =&\gamma ||Q_1(s_{t+1}, {\color{black}\eta_{t+1}},a_{t+1},\eta_{t+2},)-Q_2(s_{t+1}, {\color{black}\eta_{t+1}},a_{t+1},\eta_{t+2})||_{\infty}.
\end{align*}\normalsize
This completes the proof.
\endproof
Following the Banach fixed point theorem (see, e.g., \cite{puterman2014markov}), Theorem \ref{prop:contraction} suggests that by repeatedly updating the action value function in \eqref{eq:risk-averse-bellman}, it converges to a unique function corresponding to the optimal action value function $Q^*$ of the infinite-horizon MDP problem \eqref{ECRMs}.

Using the Bellman operator defined in Eq.\ \eqref{eq:risk-averse-bellman}, we propose a risk-averse deep Q-learning (DQN) framework in Algorithm \ref{alg:risk-averse-Q}. The risk-neutral DQN algorithm has been discussed and tested in, e.g., \cite{mnih2013playing,mnih2015human}. The differences between our risk-averse DQN and its risk-neutral counterpart are outlined as follows. First, as we augment the action space to include auxiliary variable $\eta_{t}$, we discretize the space of $\eta_t$ to have a dimension of $D$. The maximum of the right-hand side of Eq.\ \eqref{eq:cvar} is attained at $\eta_t^* = \text{VaR}_{\alpha_t}[r_t] := \inf\{v: \text{Pr}(r_t\le v)\ge \alpha_t\}$. Thus, we can set the lower and upper bounds of $\eta_t$ to be the lower and upper bounds of the immediate reward $r_t$ ($\underline{r}_t, \bar{r}_t$), respectively. Then, the $d$-th element of the $\eta_t$-space is $\underline{r}_t + \frac{\bar{r}_t-\underline{r}_t}{D}*d$ for all $d=0,1,\ldots, D$. Second, when we store the transitions into the replay memory, we not only store the state-action pair $(s_t,a_t,\eta_{t+1},s_{t+1})$, but also add the previous action $\eta_t$ to the tuple because the calculation in the risk-averse Bellman operator \eqref{eq:risk-averse-bellman} requires both the current action $\eta_{t+1}$ and the previous action $\eta_t$.

\begin{algorithm}[ht!]
\caption{Risk-averse deep Q-Learning with experience replay.}
\begin{algorithmic}[1]
\label{alg:risk-averse-Q}
  \scriptsize
\STATE Initialize replay memory $\mathcal{D}$ to capacity $N$.
\STATE Initialize action value function $Q$ with random weights $\theta$.
\STATE Initialize target action value function $\hat{Q}$ with random weights. $\hat{\theta}$
  \FOR {episode $k=1,\ldots, N$}
    \STATE Initialize $s_1$ {\color{black}and $\eta_1$}.
    \STATE With probability $\epsilon$ select a random action $(a_1,\eta_{2})$.
    \STATE Otherwise select $(a_1,\eta_2)=\arg\max_{a^{\prime},\eta^{\prime}}Q(s_1, {\color{black}\eta_{1}},a^{\prime},\eta^{\prime}; \theta)$.
    \STATE Execute action $a_1$ and observe reward $r_1$ and next state $s_{2}$.
    \FOR {time step $t=2,\ldots, T$}
    \STATE With probability $\epsilon$ select a random action $(a_t,\eta_{t+1})$.
    \STATE Otherwise select $(a_t,\eta_{t+1})=\arg\max_{a^{\prime},\eta^{\prime}}Q(s_t, {\color{black}\eta_{t}},a^{\prime},\eta^{\prime}; \theta)$.
    \STATE Execute action $a_t$ and observe reward $r_t$ and next state $s_{t+1}$.
    \STATE Store transition $(s_t, a_t, \eta_t, \eta_{t+1}, r_t, s_{t+1})$ in replay memory $\mathcal{D}$.
    \STATE Sample random minibatch of $M$ transitions  from $\mathcal{D}$: $\{(s_j, a_j, \eta_{j}, \eta_{j+1}, r_j, s_{j+1})\}_{j=1}^M$.
    \FOR {$j=1,\ldots, M$}
      \STATE Set $y_j = \begin{cases}
      -\frac{\lambda_j
      }{\alpha_j}[\eta_j-r_j]_++(1-\lambda_j)r_j,\ \text{for terminal $s_{j+1}$}, \\
      \Big(-\frac{\lambda_j
      }{\alpha_j}[\eta_j-r_j]_++(1-\lambda_j)r_j+\gamma\lambda_{j+1}\eta_{j+1}\\
      +\gamma\max_{a^{\prime},\eta^{\prime}}\hat{Q}(s_{j+1}, {\color{black}\eta_{j+1}},a^{\prime},\eta^{\prime};\hat{\theta})\Big),\ \text{otherwise}.
      \end{cases}$
      \ENDFOR
      \STATE Calculate loss $\mathcal{L}=\frac{1}{M}\sum_{j=1}^M(y_j-Q(s_j, {\color{black}\eta_{j}}, a_j, \eta_{j+1}; \theta))^2$.
      \STATE Perform a gradient descent step to minimize the loss $\mathcal{L}$ with respect to the network parameter $\theta$.
      \STATE Every $C$ steps reset $\hat{Q}=Q$.
  \ENDFOR
  \ENDFOR
\end{algorithmic}
\end{algorithm}

\section{Numerical Studies of Risk-Averse Deep Q-Learning}
We implement Algorithm \ref{alg:risk-averse-Q} based on a neural network with two hidden layers, with ReLU activation. All our experiments use \texttt{Adam} \cite{kingma2014adam} as the stochastic gradient optimizer with learning rate 0.001. The discount factor $\gamma$ is set to 0.98, and the confidence level for CVaR is set to $\alpha_t=0.05,\ \forall t\ge2$. 
We use $\epsilon$-greedy selection method \cite{sutton2018reinforcement} with a minimum $\epsilon$-value of 0.01, maximum $\epsilon$-value of 1 and decay of 0.95, respectively.
\subsection{An Illustrative 10-State MDP}
\begin{figure}[ht!]
	\centering
	\begin{subfigure}[b]{\linewidth}
	\centering
	\includegraphics[width=0.8\textwidth]{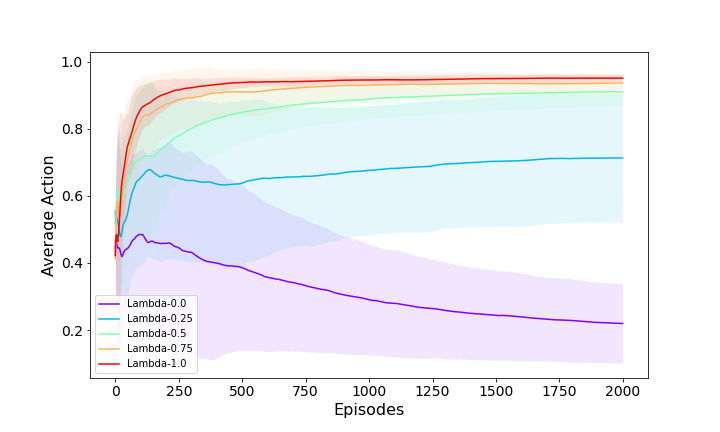}
	\caption{Average action in the 10-state MDP with varying $\lambda_t$ (where 0 represents action $a_0$ and 1 represents action $a_1$).}
	\end{subfigure}
    \begin{subfigure}[b]{\linewidth}
    \centering
    \includegraphics[width=0.8\textwidth]{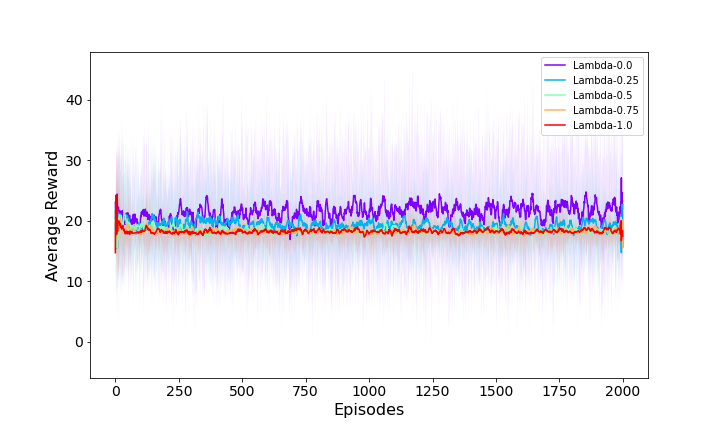}
    \caption{Average reward in the 10-state MDP with varying $\lambda_t$.}
    \end{subfigure}
    \begin{subfigure}[b]{\linewidth}
    \centering
    \includegraphics[width=0.8\textwidth]{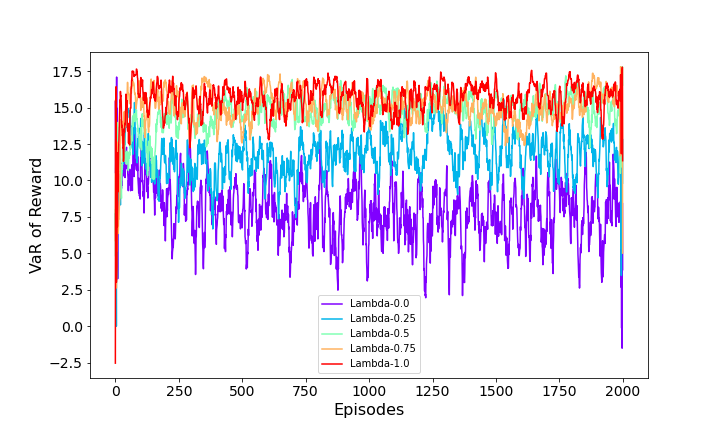}
    \caption{$\text{VaR}_{\alpha_t}$ of the reward in the 10-state MDP with varying $\lambda_t$.}
    \end{subfigure}
    \caption{Performance of risk-averse DQN on the 10-state MDP with varying $\lambda_t$.}
    \label{fig:RA-DQN-10-state}
\end{figure}

We first consider a simple 10-state MDP as an illustrative example. The MDP has ten states $s_0,s_1,\ldots, s_{9}$,  where $s_0$ is the initial state and $s_{9}$ is the terminal state. Each state has two actions $a_0$ and $a_1$. Action $a_0$ generates an immediate reward following a Normal distribution $\mathcal{N}(2.5, 4^2)$, and action $a_1$ generates an immediate reward following a Normal distribution $\mathcal{N}(2, 0.1^2)$. Clearly, action $a_0$ generates immediate rewards with a higher mean value but also higher variance. Each action always moves the state from $s_i$ to $s_{i+1}$. We train the model over 2000 episodes where each episode starts at state $s_0$ and continues until a terminal state is reached. When the terminal state is reached, a new episode restarts. All the hyperparameters are kept the same across different trials, except for the risk parameters $\lambda_t$ which is swept across 0, 0.25, 0.5, 0.75, 1. For each of these values, we perform the task over 10 independent simulation runs. 
The average action and reward over the 10 simulation runs are recorded in Figures \ref{fig:RA-DQN-10-state}(a) and \ref{fig:RA-DQN-10-state}(b), respectively, where the colored shades represent the standard deviation. We also plot the $\text{VaR}_{\alpha_t}$ of the reward with $\alpha_t=0.05$ and varying $\lambda_t$ (i.e., 5\% percentile of the reward) in Figure \ref{fig:RA-DQN-10-state}(c).


Figure \ref{fig:RA-DQN-10-state}(a) shows a clear trend that the risk-neutral DQN ($\lambda_t=0,\ \forall t\ge2$) tends to choose action $a_0$ as episode evolves, which has a higher reward mean. This can be also seen from Figures \ref{fig:RA-DQN-10-state}(b) and (c) that the risk-neutral DQN produces higher reward mean, higher reward variance and lower $\text{VaR}_{\alpha_t}$ of the reward. On the other hand, as long as we add risk-aversion to the problem by setting $\lambda_t>0$, all other four approaches tend to choose action $a_1$, which leads to lower mean, lower variance and higher $\text{VaR}_{\alpha_t}$ of the reward.

\subsection{A 7-State Random Walk MDP}
\begin{figure}[ht!]
    \centering
    \includegraphics[width=0.4\textwidth]{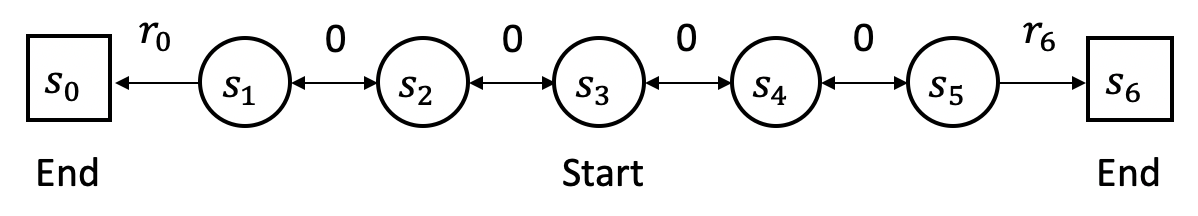}
    \caption{A random walk MDP with five (normal) states
and two terminal states $s_0$ and $s_6$. State $s_3$ is the start state. Values alongside the arrows are rewards.}
    \label{fig:randow-walk}
\end{figure}

We test our algorithm on a well-studied random walk MDP \cite{sutton1988learning, sutton2018reinforcement} with the following settings. There are 7 states $s_0,\ s_1,\ldots,s_6$ and $s_0,\ s_6$ are the terminal states (see Figure \ref{fig:randow-walk}). We start at the middle state $s_3$ and each non-terminal state has two actions that cause left- and right-transitions to neighboring states. (Here, $a_0$ represents going left, and $a_1$ represents going right.) The rewards in each transition are 0, except for transitions into the left and right terminal states $s_0, s_6$, which we denote as $r_0, r_6$, respectively. 

\begin{figure}[ht!]
	\centering
	\begin{subfigure}[b]{\linewidth}
	\centering
		\includegraphics[width=0.8\textwidth]{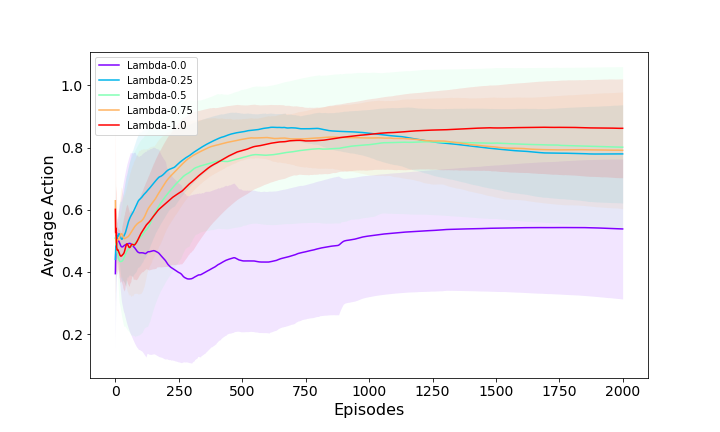}
		\caption{Average action in the 7-state random walk with varying $\lambda_t$ (where 0 represents action $a_0$ and 1 represents action $a_1$).}
	\end{subfigure}
    \begin{subfigure}[b]{\linewidth}
    \centering
        \includegraphics[width=0.8\textwidth]{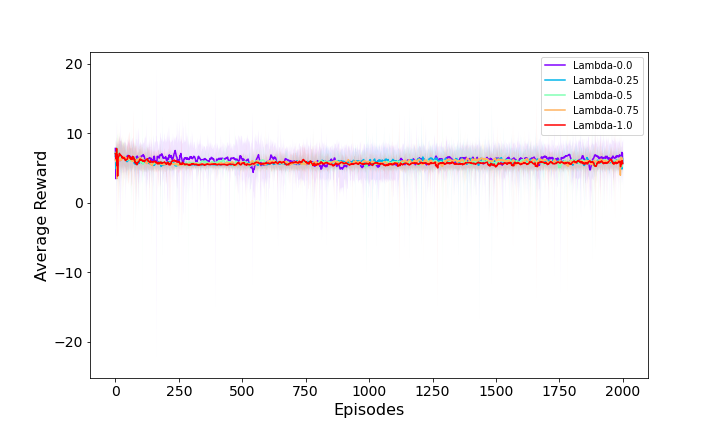}
        \caption{Average reward in the 7-state random walk with varying $\lambda_t$.}
    \end{subfigure}
        \begin{subfigure}[b]{\linewidth}
    \centering
        \includegraphics[width=0.8\textwidth]{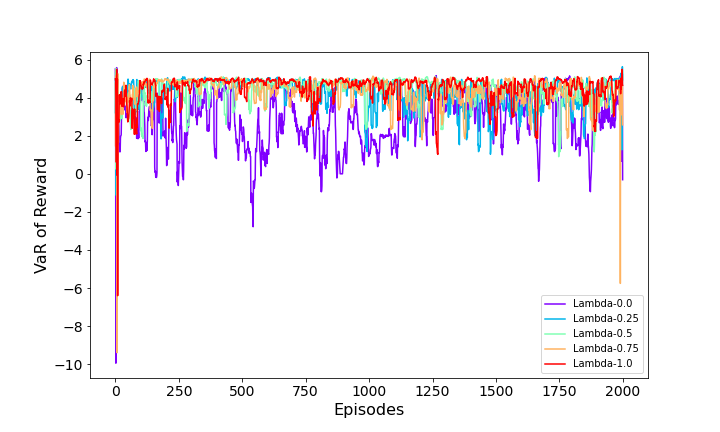}
        \caption{$\text{VaR}_{\alpha_t}$ of reward in the 7-state random walk with varying $\lambda_t$.}
    \end{subfigure}
    \caption{Performance of risk-averse DQN on the 7-state random walk with varying $\lambda_t$.}
    \label{fig:RA-DQN-random-walk}
\end{figure}

We set $r_0 = 10 - x_0,\ r_6 = 10 - x_6$ where $x_0$ follows log-normal distribution with $\mu_0 = 0.5$ and $\sigma_0 = 1$, and $x_6$ follows log-normal distribution with $\mu_6 = 1.5$ and $\sigma_6 = 0.1$. In this way, we ensure that the distributions of $r_0$ and $r_6$ are left skewed and we obtain positive rewards with high probability. Note that these parameters $\{\mu_i\}_{i=0,6},\ \{\sigma_i\}_{i=0,6}$ are the means and standard deviations of $\{x_i\}_{i=0,6}$'s natural logarithm. The means and standard deviations of the reward themselves are $\mathbb{E}[r_0]=7.28,\
\text{Var}[r_0]=12.64,\ \mathbb{E}[r_6]=5.5,\  \text{Var}[r_6]=0.2$. In this case, $r_0$ has a higher mean value, but also comes with a heavy left tail with a much higher variance.
We still fix $\alpha_t=0.05$, vary the risk parameter $\lambda_t$ from 0 to 1, and present the average action, average reward, and $\text{VaR}_{\alpha_t}$ of the reward in Figure \ref{fig:RA-DQN-random-walk}.

From Figure \ref{fig:RA-DQN-random-walk}(a), the risk-neutral DQN tends to oscillate between action $a_0$ and $a_1$ as it tends to produce an average action of around 0.5 with the highest variance, while all other four risk-averse DQN tend to choose action $a_1$. In terms of the average reward, the gaps between risk-neutral and risk-averse models are much smaller than the ones in the 10-state MDP case, mainly because the higher reward mean of action $a_0$ is offset by the possible the worst-case loss (i.e., negative reward) due to the heavy left tail. Moreover, both the average action and reward produced by risk-neutral DQN have larger variance than their risk-averse counterparts. From Figure \ref{fig:RA-DQN-random-walk}(c), the risk-averse DQN with $\lambda_t=1,\ \forall t\ge 2$ produces the highest $\text{VaR}_{\alpha_t}$ of the reward, indicating the robustness of our approach.

\section{CONCLUSIONS}
In this paper, we considered risk-averse MDPs by incorporating dynamic time-consistent risk measures. Specifically, we showed that the ECRMs are time consistent in a discounted infinite horizon setting following the definition given in \cite{ruszczynski2010risk}. Another appealing property of ECRMs is that any risk-averse MDPs can be recast as risk-neutral MDPs with augmented action space and manipulation on the immediate rewards. Furthermore, we proved that the risk-averse Bellman operator is a contraction mapping. This opened up opportunities for a wide range of algorithms, and we extended the risk-neutral DQN to a risk-averse variant. Our numerical experiments revealed the benefits of our models as compared to risk-neutral approaches, because  they reduced the variance and increased the robustness of the results.

All configurations in this paper used constant risk parameters across all episodes, and we leave the adaptive setting where the risk parameters change over time for future research. Another interesting research direction is to apply the risk measure used in this paper to a distributionally robust RL scheme and to prove that the associated risk-averse distributional Bellman operator is a contraction mapping.






\section*{ACKNOWLEDGMENT}
We thank Dr.\ Lei Ying at the University of Michigan for constructive feedback and Ms.\ Chunan Huang for setting up the MDP environment. The authors acknowledge the support from the US Department of Energy grant \#DE-SC0018018.


\bibliographystyle{IEEEtran}
\bibliography{Xian_bib}

\begin{thebibliography}{10}
\providecommand{\url}[1]{#1}
\csname url@samestyle\endcsname
\providecommand{\newblock}{\relax}
\providecommand{\bibinfo}[2]{#2}
\providecommand{\BIBentrySTDinterwordspacing}{\spaceskip=0pt\relax}
\providecommand{\BIBentryALTinterwordstretchfactor}{4}
\providecommand{\BIBentryALTinterwordspacing}{\spaceskip=\fontdimen2\font plus
\BIBentryALTinterwordstretchfactor\fontdimen3\font minus
  \fontdimen4\font\relax}
\providecommand{\BIBforeignlanguage}[2]{{%
\expandafter\ifx\csname l@#1\endcsname\relax
\typeout{** WARNING: IEEEtran.bst: No hyphenation pattern has been}%
\typeout{** loaded for the language `#1'. Using the pattern for}%
\typeout{** the default language instead.}%
\else
\language=\csname l@#1\endcsname
\fi
#2}}
\providecommand{\BIBdecl}{\relax}
\BIBdecl

\bibitem{puterman2014markov}
M.~L. Puterman, \emph{{Markov Decision Processes: Discrete Stochastic Dynamic
  Programming}}.\hskip 1em plus 0.5em minus 0.4em\relax John Wiley \& Sons,
  2014.

\bibitem{nilim2005robust}
A.~Nilim and L.~El~Ghaoui, ``Robust control of {M}arkov decision processes with
  uncertain transition matrices,'' \emph{Operations Research}, vol.~53, no.~5,
  pp. 780--798, 2005.

\bibitem{iyengar2005robust}
G.~N. Iyengar, ``Robust dynamic programming,'' \emph{Mathematics of Operations
  Research}, vol.~30, no.~2, pp. 257--280, 2005.

\bibitem{wiesemann2013robust}
W.~Wiesemann, D.~Kuhn, and B.~Rustem, ``Robust {M}arkov decision processes,''
  \emph{Mathematics of Operations Research}, vol.~38, no.~1, pp. 153--183,
  2013.

\bibitem{lim2013reinforcement}
S.~H. Lim, H.~Xu, and S.~Mannor, ``Reinforcement learning in robust markov
  decision processes,'' \emph{Advances in Neural Information Processing
  Systems}, vol.~26, 2013.

\bibitem{xu2010distributionally}
H.~Xu and S.~Mannor, ``Distributionally robust {M}arkov decision processes,''
  \emph{Advances in Neural Information Processing Systems}, vol.~23, 2010.

\bibitem{yu2015distributionally}
P.~Yu and H.~Xu, ``Distributionally robust counterpart in {M}arkov decision
  processes,'' \emph{IEEE Transactions on Automatic Control}, vol.~61, no.~9,
  pp. 2538--2543, 2015.

\bibitem{nakao2021distributionally}
H.~Nakao, R.~Jiang, and S.~Shen, ``Distributionally robust partially observable
  markov decision process with moment-based ambiguity,'' \emph{SIAM Journal on
  Optimization}, vol.~31, no.~1, pp. 461--488, 2021.

\bibitem{heger1994consideration}
M.~Heger, ``Consideration of risk in reinforcement learning,'' in \emph{Machine
  Learning Proceedings 1994}.\hskip 1em plus 0.5em minus 0.4em\relax Elsevier,
  1994, pp. 105--111.

\bibitem{coraluppi2000mixed}
S.~P. Coraluppi and S.~I. Marcus, ``Mixed risk-neutral/minimax control of
  discrete-time, finite-state markov decision processes,'' \emph{IEEE
  Transactions on Automatic Control}, vol.~45, no.~3, pp. 528--532, 2000.

\bibitem{howard1972risk}
R.~A. Howard and J.~E. Matheson, ``Risk-sensitive {M}arkov decision
  processes,'' \emph{Management Science}, vol.~18, no.~7, pp. 356--369, 1972.

\bibitem{markowitz2000mean}
H.~M. Markowitz and G.~P. Todd, \emph{Mean-variance analysis in portfolio
  choice and capital markets}.\hskip 1em plus 0.5em minus 0.4em\relax John
  Wiley \& Sons, 2000, vol.~66.

\bibitem{borkar2002q}
V.~S. Borkar, ``Q-learning for risk-sensitive control,'' \emph{Mathematics of
  Operations Research}, vol.~27, no.~2, pp. 294--311, 2002.

\bibitem{tamar2012policy}
A.~Tamar, D.~Di~Castro, and S.~Mannor, ``Policy gradients with variance related
  risk criteria,'' in \emph{Proceedings of the 29th International Coference on
  International Conference on Machine Learning}, 2012, pp. 1651--1658.

\bibitem{la2013actor}
P.~La and M.~Ghavamzadeh, ``Actor-critic algorithms for risk-sensitive
  {MDPs},'' \emph{Advances in neural information processing systems}, vol.~26,
  2013.

\bibitem{artzner1999coherent}
P.~Artzner, F.~Delbaen, J.-M. Eber, and D.~Heath, ``Coherent measures of
  risk,'' \emph{Mathematical Finance}, vol.~9, no.~3, pp. 203--228, 1999.

\bibitem{rockafellar2000optimization}
R.~T. Rockafellar, S.~Uryasev \emph{et~al.}, ``Optimization of conditional
  value-at-risk,'' \emph{Journal of Risk}, vol.~2, no.~3, pp. 21--42, 2000.

\bibitem{rockafellar2002conditional}
R.~T. Rockafellar and S.~Uryasev, ``Conditional value-at-risk for general loss
  distributions,'' \emph{Journal of Banking \& Finance}, vol.~26, no.~7, pp.
  1443--1471, 2002.

\bibitem{ruszczynski2006optimization}
A.~Ruszczy{\'n}ski and A.~Shapiro, ``Optimization of convex risk functions,''
  \emph{Mathematics of Operations Research}, vol.~31, no.~3, pp. 433--452,
  2006.

\bibitem{shapiro2009lectures}
A.~Shapiro, D.~Dentcheva, and A.~Ruszczy{\'n}ski, \emph{Lectures on Stochastic
  Programming: {M}odeling and {T}heory}.\hskip 1em plus 0.5em minus 0.4em\relax
  SIAM, 2009.

\bibitem{yu2017dynamic}
P.~Yu, W.~B. Haskell, and H.~Xu, ``Dynamic programming for risk-aware
  sequential optimization,'' in \emph{2017 IEEE 56th Annual Conference on
  Decision and Control (CDC)}.\hskip 1em plus 0.5em minus 0.4em\relax IEEE,
  2017, pp. 4934--4939.

\bibitem{chow2014algorithms}
Y.~Chow and M.~Ghavamzadeh, ``Algorithms for {CVaR optimization in MDPs},''
  \emph{Advances in Neural Information Processing Systems}, vol.~27, 2014.

\bibitem{tamar2015optimizing}
A.~Tamar, Y.~Glassner, and S.~Mannor, ``Optimizing the {CVaR} via sampling,''
  in \emph{Twenty-Ninth AAAI Conference on Artificial Intelligence}, 2015.

\bibitem{rudloff2014time}
B.~Rudloff, A.~Street, and D.~M. Vallad{\~a}o, ``Time consistency and risk
  averse dynamic decision models: {D}efinition, interpretation and practical
  consequences,'' \emph{European Journal of Operational Research}, vol. 234,
  no.~3, pp. 743--750, 2014.

\bibitem{ruszczynski2010risk}
A.~Ruszczy{\'n}ski, ``Risk-averse dynamic programming for {M}arkov decision
  processes,'' \emph{Mathematical Programming}, vol. 125, no.~2, pp. 235--261,
  2010.

\bibitem{chow2013stochastic}
Y.-L. Chow and M.~Pavone, ``Stochastic optimal control with dynamic,
  time-consistent risk constraints,'' in \emph{2013 American Control
  Conference}.\hskip 1em plus 0.5em minus 0.4em\relax IEEE, 2013, pp. 390--395.

\bibitem{chow2014framework}
------, ``A framework for time-consistent, risk-averse model predictive
  control: Theory and algorithms,'' in \emph{2014 American Control
  Conference}.\hskip 1em plus 0.5em minus 0.4em\relax IEEE, 2014, pp.
  4204--4211.

\bibitem{homem2016risk}
T.~Homem-de Mello and B.~K. Pagnoncelli, ``Risk aversion in multistage
  stochastic programming: A modeling and algorithmic perspective,''
  \emph{European Journal of Operational Research}, vol. 249, no.~1, pp.
  188--199, 2016.

\bibitem{sutton1988learning}
R.~S. Sutton, ``Learning to predict by the methods of temporal differences,''
  \emph{Machine Learning}, vol.~3, no.~1, pp. 9--44, 1988.

\bibitem{sutton2018reinforcement}
R.~S. Sutton and A.~G. Barto, \emph{{Reinforcement Learning: An
  Introduction}}.\hskip 1em plus 0.5em minus 0.4em\relax MIT Press, 2018.

\bibitem{bellman1966dynamic}
R.~Bellman, ``Dynamic programming,'' \emph{Science}, vol. 153, no. 3731, pp.
  34--37, 1966.

\bibitem{mnih2013playing}
V.~Mnih, K.~Kavukcuoglu, D.~Silver, A.~Graves, I.~Antonoglou, D.~Wierstra, and
  M.~Riedmiller, ``Playing atari with deep reinforcement learning,''
  \emph{arXiv preprint arXiv:1312.5602}, 2013.

\bibitem{mnih2015human}
V.~Mnih, K.~Kavukcuoglu, D.~Silver, A.~A. Rusu, J.~Veness, M.~G. Bellemare,
  A.~Graves, M.~Riedmiller, A.~K. Fidjeland, G.~Ostrovski \emph{et~al.},
  ``Human-level control through deep reinforcement learning,'' \emph{Nature},
  vol. 518, no. 7540, pp. 529--533, 2015.

\bibitem{kingma2014adam}
D.~P. Kingma and J.~Ba, ``Adam: {A} method for stochastic optimization,''
  \emph{arXiv preprint arXiv:1412.6980}, 2014.

\end{thebibliography}

\end{document}